# Anytime Algorithms for Speech Parsing?*

Günther Görz    Marcus Kesseler

University of Erlangen-Nürnberg, IMMD VIII

goerz@informatik.uni-erlangen.de

TOPICAL PAPER

## Abstract

This paper discusses to which extent the concept of "anytime algorithms" can be applied to parsing algorithms with feature unification. We first try to give a more precise definition of what an anytime algorithm is. We argue that parsing algorithms have to be classified as contract algorithms as opposed to (truly) interruptible algorithms. With the restriction that the transaction being active at the time an interrupt is issued has to be completed before the interrupt can be executed, it is possible to provide a parser with limited anytime behavior, which is in fact being realized in our research prototype.

## 1 Introduction

The idea of *"anytime algorithms"*, which has been around in the field of planning for some time[1], has recently been suggested for application in natural language and speech processing (NL/SP)[2]. An anytime algorithm is an algorithm "whose quality of results degrades gracefully as computation time decreases" ([Russell and Zilberstein 1991], p. 212). In the following we will first give a more specific definition of which properties allow an algorithm to be implemented and used as an anytime algorithm. We then apply this knowledge to a specific aspect of NL/SP, namely parsing algorithms in a speech understanding system. In the Appendix we present the APC protocol which supports anytime computations.

We will discuss these matters in the framework of the Verbmobil joint research project[3], where we are working on the implementation of an incremental chart parser[4]. The conception of this parser has been derived from earlier work by the first author[5].

---

[1] cf. e.g. [Russell and Zilberstein 1991]

[2] so [Wahlster 1992] in his invited talk at COLING-92

[3] The Verbmobil joint research project has been defined in the document [Verbmobil Report 1991]

[4] the Verbmobil/15 parser, cf. [Weber 1992]

[5] the GuLP parser, cf. [Görz 1988].

## 2 Anytime Algorithms

[Dean and Boddy 1988] give the following characterization of anytime algorithms:

1. they lend themselves to preemptive scheduling techniques (i.e., they can be suspended and resumed with negligible overhead),

2. they can be terminated at any time and will return some answer, and

3. the answers returned improve in some well-behaved manner as a function of time.

Unfortunately this characterization does not make a clear distinction between the implementation of an algorithm and the algorithm as such.

Point (1) is true of a great many algorithms implemented on preemptive operating systems.

Point (2) can be made true for any algorithm by adding an explicit `Result` slot, that is preset by a value denoting a *void result*. Let us call the implementation of an anytime algorithm an *anytime producer*. Accordingly we name the entity interested in the result of such an anytime computation the *anytime consumer*. Figure 1 shows two such processes in a tightly coupled synchronization loop. Figure 2 shows the same communicating processes decoupled by the introduction of the `Result` slot. Note that synchronisation is much cheaper in terms of perceived complexity for the programmer and runtime synchronisation overhead (just the time to check and eventually traverse the mutual exclusion barrier). In such an architecture producer and consumer work under a regime that allows the consumer to interrupt the producer at *any time* and demand a result. The risk that the consumer incurs by such flexibility is a certain non-zero probability that this result is void[6] or unchanged since the last result retrieval.

---

[6] The failure to provide an answer within a given amount of time may in itself be an interesting and meaningful result for the anytime consumer.

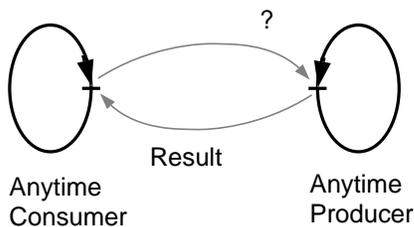

Figure 1: Tightly coupled processes with complex synchronization internals.

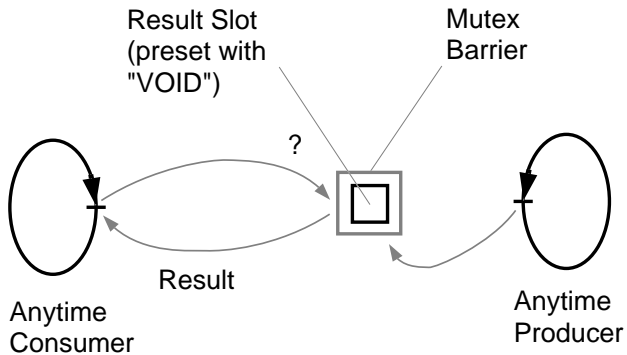

Figure 2: Processes decoupled by using a result slot protected by a simple mutual exclusion barrier.

Point (3) is surely a much too strong restriction, since it is not always possible to define what exactly an *improvement* is for any given algorithm. In NL/SP, where we are often dealing with scored hypotheses, it is difficult, if not impossible, to devise algorithms that supply answers that improve monotonically as a function of invested computational resources (time or processing units in a parallel architecture).

We propose the following characterization of anytime algorithms:

> An algorithm is fit to be used as an anytime producer if its implementation yields a program that has a *Result Production Granularity (RPG)* that is compatible with the time constraints of the consumer.

The notion of RPG is based on the following observation: Computations being performed on finite state machines do not proceed directly from goal state to goal state. Instead they go through arbitrarily large sequences of states that yield no extractable or intelligible data to an outside observer. To interrupt a producer on any of these intermediate states is fruitless, since the result obtained could at best, according to the observation made on point (2) above, be the result that was available in the last goal state of the producer. From the point of view of the consumer the *transitions* from goal state to goal state in the producer are *atomic transactions*.

The average length of these transactions in the algorithm correspond to average time intervals in the implementation, so that we can speak of a *granularity* with which results are produced.

The time constraints under which the consumer is operating then give the final verdict if the implementation of an algorithm is usable as an anytime producer. Let us illustrate this by an example: In a real-time NL/SP-system the upper bound for the RPG will typically be in the range of 10–100ms. That is, a producer implemented with such an RPG offers the consumer the chance to trade a 500ms delay for 5 to 50 further potential solutions.

Note that goal states can also be associated with *intermediate* results in the producer algorithm. Conceptually there really is not much of a difference between a result and an intermediate result, but in highly optimized implementations there might be the need to explicitly export such intermediate results, due to data representation incompatibilities or simply because the data might be overwritten by other (non-result) data. Section 4 gives an example of how the RPG of an implementation can be reduced by identifying intermediate goal states that yield information which is of interest to the consumer.

## 3 Breadth and Depth of Analysis

In the following we will ask whether and how the idea of anytime producers can be applied within the active chart parsing algorithm scheme with feature unification. Although the analogy to decision making in planning — where the idea of anytime algorithms has been developed — seems to be rather shallow, we can, for the operation of the parser, distinguish between *depth* and *breadth* of analysis[7].

- We define *depth* of analysis as the concept refering to the growing size of information content in a feature structure over a given set of non-competing word hypotheses in a certain time segment during its computation. Larger depth corresponds to a more detailed linguistic description of the *same* objects.

- In contrast, we understand by *breadth* of analysis the consideration of linguistic descriptions resulting from the analysis of growing sets of word hypotheses, either from growing segments of the utterance to be parsed or from a larger number of competing word hypotheses in a given time segment.

To regard breadth of analysis as a measure in the context of the anytime algorithm concept is in a sense trivial: Considering only one parse, the more processing time the parser is given the larger the analyzed segment of the input utterance will be. In general, larger breadth corresponds to more information about competing word hypotheses in an (half-) open time interval as opposed to more information about a given word sequence. So, obviously, breadth of analysis does not correspond to what is intended by the concept of anytime algorithms, whereas depth of analysis meets the intention.

---

[7]not to be confused with depth-first or breadth-first search.

If an utterance is syntactically ambiguous, we can compute more parses the more processing time the parser is given. Therefore, this case is a particular instance of depth of analysis, because the same word sequence is considered, and not of breadth of analysis given the definition above. In this case one would like to get the best analysis in terms of the quality scores of its constituents first, and other readings later, ordered by score. If the parser works incrementally, what happens to be the case for the Verbmobil/15 parser[8], the intended effect can be achieved by the adjustment of a strategy parameter — namely to report the analysis of a grammatical fragment of the input utterance as soon as it is found.

At least one distinction might be useful for the Verbmobil/15 parser. In our parser a category check is performed on two chart edges for efficiency reasons, and only if this check is successful, the unification of the associated feature structures is performed. Hence, an interrupt would be admissible after the category check. In this case we emphasize a factorization of the set of constraints in two distinct subsets: phrasal constraints which are processed by the active chart parsing algorithm schema (with polynomial complexity), and functional constraints which are solved by the unification algorithm (with exponential complexity). The interface between both types of constraints is a crucial place for the introduction of control in the parsing process in general[9]

Since we use a constraint-based grammar formalism, whose central operation is the unification of feature structures, it does not make sense to admit interrupts at any time. Instead, the operation of the parser consists of a sequence of transactions. At the most coarse grained level, a transaction would be an application of the fundamental rule of active chart parsing, i.e. a series of operations which ends when a new edge is introduced into the chart, including the computation of the feature structure associated with it. Of course this argument holds when an application of the fundamental rule results in another application of it on subunits due to the recursive structure of the grammar rules[10]. Certainly one might ask whether a smaller grain size makes sense, i.e. the construction of a feature structure should itself be interruptible. In this case one could think of the possibility of an interrupt after one feature in one of the two feature structures to be unified has been processed. We think that this possibility should be rejected, since feature structures usually contain coreferences. If we consider a partial feature structure — as in an intermediate step in the unification of two feature structures — in the situation where just one feature has been processed, this structure might not be a realistic partial description of the part of speech under consideration, but simply inadequate as long as not all embedded coreferences have been established. It seems obvious that the grain size cannot be meaningfully decreased below the processing of one feature. Therefore we decided that transactions must be defined in terms of computations of whole feature structures.

Nevertheless, a possibility for interrupting the computation of a feature structure could be considered in case the set of features is divided in two classes: features which are obligatory and features which are optional. Members of the last group are candidates for constraint relaxation which seems to be relevant with respect to robustness — at least in the case of speech parsing. We have just started to work on the constraint relaxation problem, but there is no doubt that this is an important issue for further research. Nevertheless, at the time being we doubt whether the above mentioned problem with coreferences could be avoided in this case.

A further opportunity for interrupts comes up in cases where the processing of alternatives in unifying disjunctive feature structures is delayed. In this case, unification with one of the disjuncts can be considered as a transaction.

Another chance for the implementation of anytime behavior in parsing arises if we consider the grammar from a linguistic perspective as opposed to the purely formal view taken above. Since semantic construction is done by our grammar as well, the functional constraints contain a distinct subset for the purpose of semantic construction. In a separate investigation [Fischer 1994] implemented a version of $\lambda$-DRT [Pinkal 1993] within the same feature unification formalism which builds semantic structures within the framework of Discourse Representation Theory. It has been shown that the process of DRS construction can be split in two types of transactions, one which can be performed incrementally — basically the construction of event representations without temporal information — and another one which cannot be concluded before the end of an utterance has been reached — supplying temporal information. Since the first kind of transactions represents meaningful partial semantic analyses those can be supplied immediately on demand under an anytime regime.

The possibility to process interrupts with the restriction that the currently active transaction has to be completed in advance has been built into the Verbmobil/15 parser, using the APC protocol (cf. Appendix). It therefore exhibits a limited anytime behavior.

---

[8] and for GuLP as well
[9] cf. [Maxwell and Kaplan 1994]
[10] This has been implemented in the interrupt system of GuLP [Görz 1988].

## 4 Feature Unification as an Anytime Algorithm?

Up to now, in our discussion of an appropriate grain size for the unification of feature structures we considered two cases: the unification of two whole feature structures or the unification of parts of two feature structures on the level of disjuncts or individual features. In all of these cases unification is considered as a single step, neglecting its real cost, i.e. time constraints would only affect the number of unification steps, but not the execution of a particular unification operation. Alternatively, one might consider the unification algorithm itself as an anytime algorithm with a property which one might call "shallow unification". A shallow unification process would quickly come up with a first, incomplete and only partially correct solution which then, given more computation time, would have to be refined and possibly revised. It seems that this property cannot be achieved by a modification of existing unification algorithms, but would require a radically different approach. A prerequisite for that would be a sort of quality measure[11] for different partial feature structures describing a given linguistic object which is distinct from the subsumption relation. To our knowledge, the definition of such a measure is an open research question.

## 5 Conclusion

According to [Russell and Zilberstein 1991] parsing algorithms with feature unification have to be classified as contract algorithms as opposed to (truly) interruptible algorithms: They must be given a particular time allocation in advance, because interrupted at any time shorter than the contract time they will not yield useful results. At least the transaction which is active at the time an interrupt occurs has to be completed before the interrupt can be executed. With this restriction, it is possible to provide a parser with limited anytime behavior, which is in fact being realized in the current version of the Verbmobil/15 parser.

**Acknowledgements.** The authors would like to thank Gerhard Kraetzschmar, Herbert Stoyan, and Hans Weber for valuable comments on a previous version of this paper.

## Appendix: A Protocol for Anytime Producer/Consumer Processes

In the following we introduce the *APC (Anytime Producer Consumer)* protocol which allows for easy establishment of anytime producer/consumer relationships on parallel architectures.

Let `Producer` be the function that implements the producer algorithm. In a purely sequential procedural call/return implementation this function would have a control structure similar to:

```
(defun Producer (...)
  (Initialize)
  (let ((Result nil))
    (while (not (GoodEnough? Result))
      (ImproveResult))
    Result))
```

---

[11] c.f. [Russell and Wefald 1989]

The RPG of `Producer` is at least that of the function `ImproveResult`. It is finer if `ImproveResult` is itself made of loops that produce intermediate results that are exportable to consumers.

The consumer is implemented as the function `Consumer`, that at some point calls the producer:

```
(defun Consumer (...)
  .
  .
  (Producer ...)
  .
  .
)
```

We now translate `Producer` and `Consumer` into parallel processes using the APC protocol, which is directly implemented by functions that act as interfaces to the underlying communication/synchronization system. All functions implementing the protocol have the prefix `APC:` (In our implementation all of them are in the Common-Lisp package `anytime-producer-consumer`).

```
(defun AnytimeProducer (...)
  (Initialize)
  (let ((Result nil))
    (while (not (GoodEnough? Result))
      (ImproveResult)
      ;; Make Result available to consumers
      (APC:SetResult! Result)
      ;; Check for messages/instructions
      ;; from Consumer
      (APC:CheckStatus)
    Result))
```

In a parallel implementation it is not sufficient for the consumer to simply *call* the producer. The producer has to be *spawned* or *forked* as a separate process:

```
(defun AnytimeConsumer (...)
  .
  .
  ; Create a new process
  (let ((P-AnytimeProducer-1
         (APC:StartProcess (AnytimeProducer ...))))
    .
    .
    (let ((Result
           (APC:GetResult P-AnytimeProducer-1)))
      (while (not (ConsumerGoodEnough? Result))
        .
        ; Do something else, like going to sleep
        ; to give the producer some more time
        .
        (setf Result
          (APC:GetResult P-AnytimeProducer-1))
    ) )
    (APC:AbortProcess P-AnytimeProducer-1))
  .
  .
)
```

### The APC Protocol

`APC:StartProcess F` – starts a new process in which the procedure `F` is executed. This function is also responsible for the creation of the protected `Result` slot. `APC:StartProcess` returns the id of the new process.

Note that an arbitrary number of producers may be started by a consumer. A producer may of course also start other producers.

`APC:AbortProcess Proc` – aborts the process `Proc`.

`APC:SetResult! R` – sets the value of the `Result` slot to `R`.

`APC:GetResult P` – retrieves the current value of the `Result` slot from process `P`. Remember that `APC:SetResult!` and `APC:GetResult` avoid read/write conflicts by a locking mechanism that implements mutual exclusion.

`APC:ResetProcess Proc I` – restarts the process `Proc` with new input `I`.

`APC:CheckStatus [Proc]` – check if any messages or instructions have arrived from `Proc`. Often parallel software environments offer only very crude process scheduling and control primitives. The user may have to implement some of them by himself. `APC:ResetProcess`, for example, is difficult to formulate in a general way. `Reset` can also involve maintenance or cleanup work, which is clearly beyond any process-oriented implementation of `Reset`. The idea is that these user implemented control procedures are hooked into `APC:CheckStatus [Proc]`. To attain a fine-grained control relationship between consumer and producer, the user simply inserts `APC:CheckStatus` at key-positions in the code.

The APC protocol has been implemented and tested under a coarse grained parallel Common-Lisp System running on a four processor SUN-SPARC MP-670. UNIX IPC[12] shared memory and semaphores are used to implement the low-level communication and synchronisation facilities. We are currently porting the system to Solaris 2.3, with PVM (Parallel Virtual Machine, see [Dongarra, Geist, Manchek and Sundaram 1993]) as the basic communication facility. PVM would allow us to move our parallel system from the current high communication and low memory bandwidth implementation on a shared memory machine, to a low communication/high memory bandwidth implementation running on a cluster of workstations.

---

[12]Interprocess Communication Facilities